\documentclass{ieeeaccess}
\usepackage{cite}
\usepackage{amsmath,amssymb,amsfonts}
\usepackage{algorithmic}
\usepackage{graphicx}
\usepackage{textcomp}
\usepackage{caption}
\def\BibTeX{{\rm B\kern-.05em{\sc i\kern-.025em b}\kern-.08em
    T\kern-.1667em\lower.7ex\hbox{E}\kern-.125emX}}
\begin{document}
\history{Date of publication xxxx 00, 0000, date of current version xxxx 00, 0000.}
\doi{10.1109/ACCESS.2017.DOI}

\title{A Reconfigurable Fabric-Based Pneumatic Actuator with Button-Fastened Constraint Modules for Multi-Mode Actuation}
\author{\uppercase{Kentaro Kimura}\authorrefmark{1},\uppercase{Hiroki Ishizuka}\authorrefmark{1}, \uppercase{Yusuke Sakaue}\authorrefmark{1}, \uppercase{Sei Ikeda}\authorrefmark{1}, \IEEEmembership{Member, IEEE}, and \uppercase{Osamu Oshiro}\authorrefmark{1}
}
\address[1]{ Graduate School of Engineering Science, Osaka University, Osaka 560-0043, Japan (e-mail: kimura@bpe.es.osaka-u.ac.jp)}
\tfootnote{This study was supported by JST PRESTO Grant Number JPMJPR22S2 and JSPS KAKENHI Grant numbers 22H01447 and 23H01379, Japan.}

\markboth
{Kimura \headeretal: A Reconfigurable Fabric-Based Pneumatic Actuator}
{Kimura \headeretal: A Reconfigurable Fabric-Based Pneumatic Actuator}

\corresp{Corresponding author: Hiroki Ishizuka (e-mail: super.stream.1440@gmail.com).}

\begin{abstract}
Soft pneumatic actuators are widely used in soft robotics; however, their actuation modes are typically fixed by internal chamber designs or reinforcements, making post-fabrication reconfiguration difficult. This paper presents a reconfigurable fabric-based pneumatic actuator that enables rapid, tool-free, and reusable switching among multiple actuation modes by attaching constraint fabric modules to a buttoned textile sleeve. By rearranging low-stretch constraint parts on a single actuator body, the proposed system realizes four actuation modes—isotropic expansion, contraction, bending, and twisting—without redesigning or disassembling the actuator. We further introduce an attachable capacitive fabric-based tactile sensor module that can be mounted at arbitrary locations on the actuator surface for task-dependent contact sensing. Experiments demonstrated an elongation ratio of 1.2, a maximum bending angle of approximately 60~degrees, and a twisting angle of approximately 90~degrees at 24~kPa, while achieving blocking forces exceeding 10~N across all modes. The tactile sensor exhibited a monotonic, force-dependent increase in capacitance with low hysteresis (approximately 12\%) and stable cyclic responses under preload. Finally, we constructed soft robotic arms composed of multiple actuators and demonstrated in-situ motion reconfiguration for grasping by changing the attached constraint patterns. These results indicate that the proposed modular constraint-and-sensing framework improves the functional adaptability of fabric-based soft robots while preserving a simple, robust actuator structure.
\end{abstract}


\titlepgskip=-15pt

\maketitle
\section{Introduction}
\label{sec:introduction}
\PARstart{S}{o}{f}{t} robots have gained attention for their compliance and safety in physical interactions.
Made from flexible elastomers, soft robots can adapt to objects and environments via large deformations.
Unlike rigid robots, they conform to objects and interact safely without excess force.
Owing to these features, soft robotic systems have been actively investigated for applications in medical robotics, disaster response, and physical assistance~\cite{b5, b6, b17, b24}.

Despite their versatility, the actuation and target tasks of many robots, including soft robots, are largely constrained by mechanisms and structures designed at the outset.
Thus, robots often need to be designed specifically for the intended task, and they may not be able to adapt their behavior to changing requirements or situations~\cite{b1, b7, b8}.
A promising approach to address this demand is to enable robots to reconfigure their shape or structure in response to the environment and use case, thereby achieving greater adaptability to tasks and surroundings~\cite{b32}.

Modular robotic systems have long been studied as a practical way to realize such reconfigurability~\cite{b33, b34}.
By rearranging constituent elements, a robot can be transformed into a different robot with distinct capabilities.
For example, modular systems have been demonstrated that can change their morphology from a quadruped to a snake-like robot~\cite{b35}.
In soft robotics, reconfigurable robots have also been explored.
Kwok \textit{et al.} demonstrated that multiple configurations, such as a common gripper and a quadruped robot, can be constructed by reconnecting components via magnetic coupling embedded in silicone rubber~\cite{b36}.
Kang \textit{et al.} showed a flexible electronics system capable of shape reconfiguration using self-healing polymers and liquid-metal wiring~\cite{b37}.
Legrand \textit{et al.} demonstrated that simulated reconfigurable robots can be implemented as physical systems by leveraging self-healing polymers~\cite{b38}.
While these studies successfully alter robot functions through reconfiguration, they often involve rearranging the overall robot structure rather than reconfiguring the actuator itself.

Soft actuators are key mechanical elements that determine soft robot behaviors; however, attempts to reconfigure the actuators themselves remain limited~\cite{b5, b6, b17}.
Soft pneumatic actuators typically provide limited deformation modes, set during fabrication by their geometry and reinforcements.
To make soft actuators reconfigurable, it is essential to devise structural or material strategies, and several approaches have been proposed.
Yoshida~\textit{et al.} proposed a method to change actuation by locally melting a low-melting-point metal, but the demonstrated reconfiguration was limited to modifying a bending motion~\cite{b39}.
Bhat \textit{et al.} proposed an approach in which detachable tapes provide surface constraint patterns to modulate actuator deformation~\cite{b10}.
Koasaka~\textit{et al.} proposed a reversible method to change actuation by attaching a less-stretchable self-healing polymer layer on the surface of a soft pneumatic actuator~\cite{b40}.
However, these methods may require specialized materials and/or rework constrained regions when switching actuation modes, increasing the effort and time required for reconfiguration.
Several studies have explored reconfigurability by replacing the entire actuator or exchanging an outer skin~\cite{b10, b18, b19}.
While effective in controlled settings, these approaches still require disassembly or replacement of the actuator.
Therefore, a tool-free, reusable reconfiguration mechanism that is independent of the actuator body — i.e., enabling switching without disassembly, redesign, or replacement of the actuator — remains needed.

In this study, we propose a fabric-based pneumatic actuator with easily attachable and detachable constraints inspired by fabric patches used for clothing reinforcement.
Although elastomers offer high flexibility, their low stiffness often leads to rupture under high internal pressure; therefore, fiber-reinforced or fabric-based constraints are frequently used to improve durability and constrain deformation~\cite{b11, b12, b23, b25, b43}.
Viewing the actuator surface as a ``wearable-like'' substrate enables us to mount functional fabrics in a modular manner~\cite{b45, b46}.
In this work, we exploit fabrics with different stretchability as constraint elements to impose desired deformation patterns on demand.
By leveraging a button-like fastening structure commonly used in clothing, constraint fabrics can be attached and detached in a simple, tool-free manner, enabling rapid, reversible actuation switching.
Unlike conventional approaches that rely on internal chamber redesign or embedded reinforcements \cite{b4, b25}, the proposed method preserves the simplicity and robustness of a single actuator body while enabling repeated reconfiguration through reusable constraint modules.

Furthermore, the same attachment concept allows sensors to be mounted at arbitrary locations on the actuator surface, which is crucial for acquiring environmental and self-state information~\cite{b26, b30, b31}.
Such flexibility is particularly important because the informative contact region can vary across tasks (e.g., grasping vs. supporting), configurations, and interaction conditions, and because sensors may need to be repositioned or replaced in response to damage or changing requirements.
Accordingly, we also integrate a soft, attachable tactile sensor as a functional module on the actuator.

The main contributions of this paper are summarized as follows:
\begin{itemize}
\item We propose a reconfigurable, fabric-based pneumatic actuator and experimentally demonstrate that reusable, attachable constraint modules enable switching among desired deformation modes without redesigning or disassembling the entire actuator body.
\item We develop a soft tactile sensor module that can be attached to arbitrary locations on the actuator surface, enabling flexible placement for task-dependent sensing.
\item We construct soft robotic arms composed of multiple soft pneumatic actuators and demonstrate that their motion can be reconfigured by changing the attached constraints.
\end{itemize}

\section{Method}

\subsection{Principle of Actuator}
\begin{figure}[t!]
    \centering
    \includegraphics[width=1.0\columnwidth]{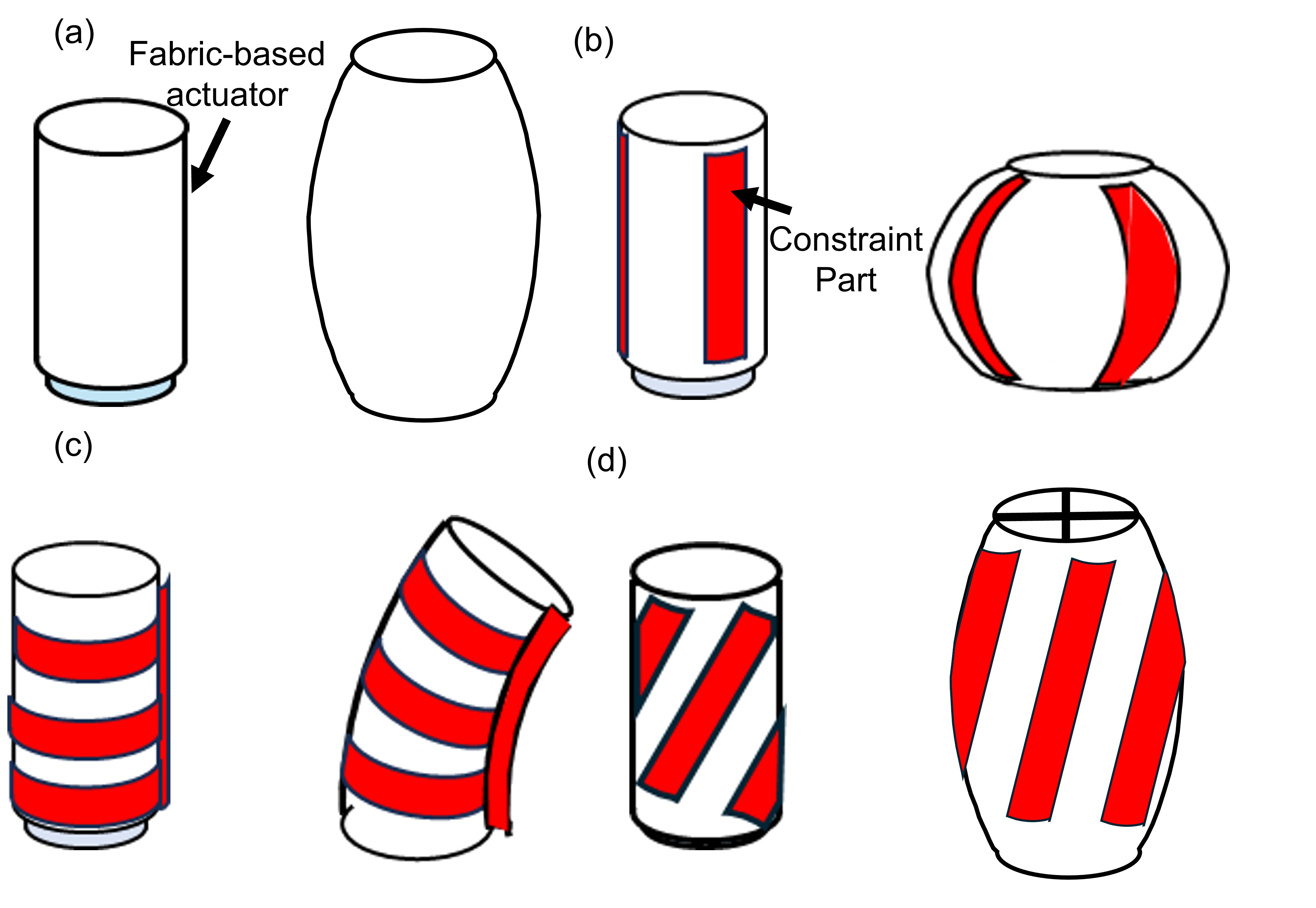} 
    \caption{Deformation of a fabric-based pneumatic actuator can be designed by taking into consideration stretchability. (a) A fabric-based pneumatic actuator expands nearly uniformly when not constrained. (b)  A fabric-based pneumatic actuator contracts when all its sides are constrained. (c) A fabric-based pneumatic actuator bends when one of its sides is constrained. (d) A fabric-based actuator twists when spiral constraints are applied.}
    \label{fig1}
\end{figure}
Figure~\ref{fig1} illustrates the concept of the proposed fabric-based pneumatic actuator. By changing the placement of externally attachable \emph{constraint parts} on the actuator surface, the deformation mode of a single actuator can be reconfigured after fabrication.

The proposed actuator has a hybrid structure comprising an elastic elastomeric bladder (an internal expansion chamber) and a stretchable fabric sleeve. Buttons are integrated on the fabric sleeve, and the constraint parts are mechanically fixed to the actuator surface via these buttons, enabling rapid attachment, detachment, and rearrangement.

The base design of the fabric-based pneumatic actuator is an isotropic cylindrical body that exhibits nearly uniform radial expansion when the internal pressure increases, as shown in Fig.~\ref{fig1}(a). By selectively attaching non-stretchable constraint parts to the sleeve, local non-uniformity in stiffness is introduced. The key mechanism of the proposed actuator is the induced strain anisotropy caused by local constraints, which converts the original uniform expansion into task-oriented deformations. Therefore, the specific deformation mode is determined by the spatial arrangement of the constraint parts.

For example, Contraction is achieved by attaching less-stretchable constraint parts around the entire circumference of the actuator, as shown in Fig.~\ref{fig1}(b). These constraint parts locally fix the lateral surface length, creating an alternating pattern of constrained and unconstrained regions along the body. Under internal pressure, each constrained segment maintains its longitudinal length, while each unconstrained segment expands predominantly in the radial direction, forming a bulged (approximately spherical) shape. The accumulation of these local radial expansions results in a net reduction in the overall axial length, thereby inducing contraction of the actuator.

When constraint parts are placed on one side of the actuator and perpendicular to the elongation direction, axial elongation on the constrained side is inhibited relative to the unconstrained side. This strain gradient produces a bending motion toward the constrained surface, as shown in Fig.~\ref{fig1}(c).

Twisting is achieved by arranging the constraint parts in a helical pattern along the lateral surface, as shown in Fig.~\ref{fig1}(d). This configuration suppresses length changes along the helical path while allowing the unconstrained regions to expand. The resulting surface deformation generates rotational displacement around the actuator axis. The twisting angle depends on the input pneumatic pressure and the constraint density, which is determined by the number of helical turns per unit length.


\subsection{Fabrication Process of Actuator}
\begin{figure}[t!]
    \centering
    \includegraphics[width=1.0\columnwidth]{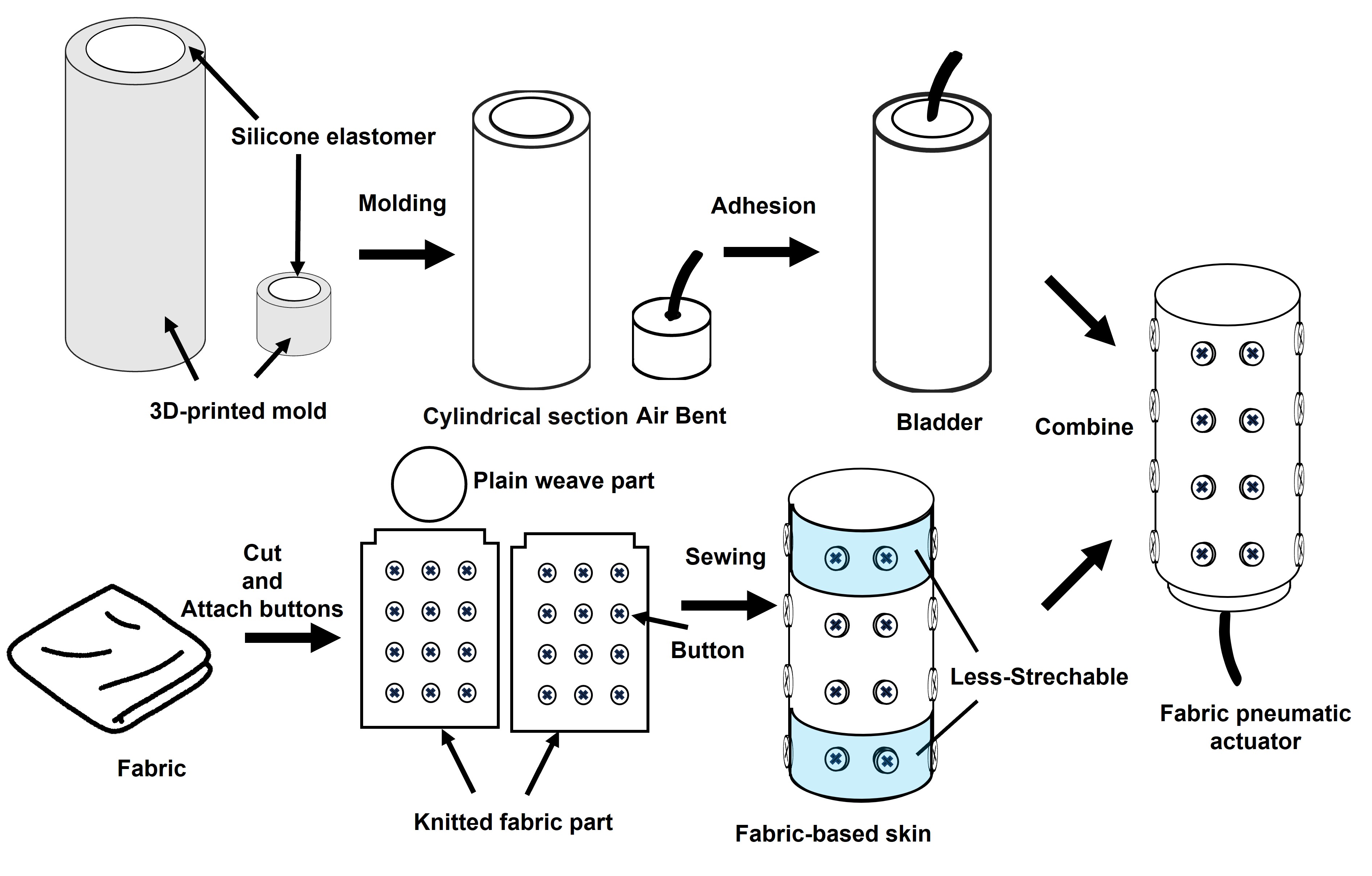} 
    \caption{Fabrication process of the fabric-based pneumatic actuator. The internal silicone rubber body for the pneumatic actuator was formed using resin molds (Upper). The external fabric skin
was a fabricated stretchable textile (Bottom). Finally, the silicone rubber body was covered with the fabric skin (Right). }
    \label{fig2}
\end{figure}
\begin{figure}[b!]
    \centering
    \includegraphics[width=1.0\columnwidth]{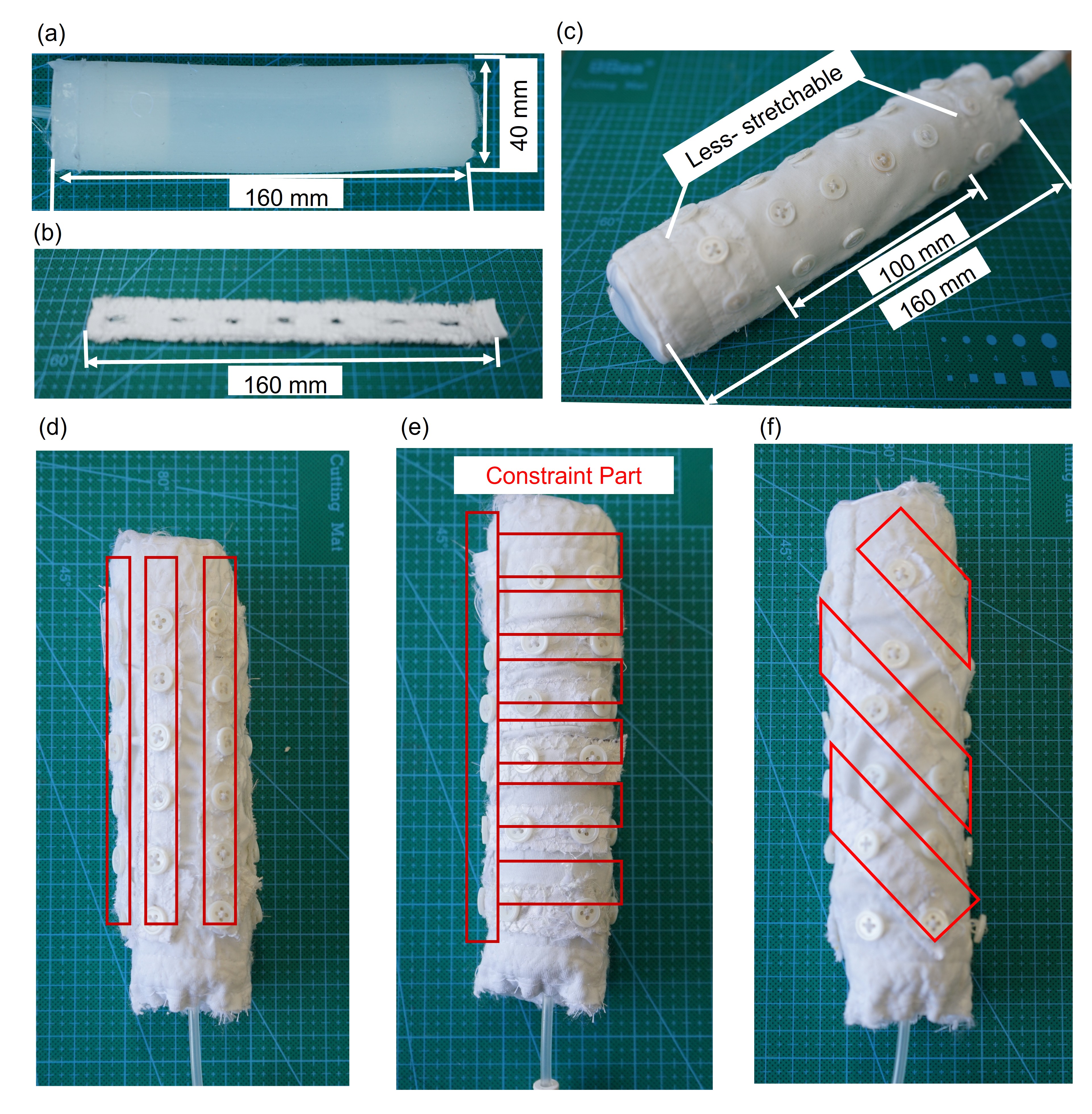} 
    \caption{Photographs of fabric-based pneumatic actuator and its components. (a) A silicone rubber bladder is used inside the actuator. (b) A non-elastic deformation constraint part.  (c) An actuator made of highly elastic fabric and a button for attaching a constraint part. (d) (e) (f) Constraint patterns for each actuator.}
    \label{fig3}
\end{figure}
The fabrication of the proposed fabric-based pneumatic actuator integrates silicone casting with textile assembly, as shown in Fig.~\ref{fig2}. The process is categorized into the fabrication of the internal silicone rubber body for pneumatic actuation (Fig.~\ref{fig2} (Upper)) and the fabric skin  (Fig.~\ref{fig2} (Bottom)).
The internal silicone rubber body was constructed by assembling multiple silicone rubber components. The cylindrical section and the air inlet section were fabricated using soft silicone rubber (Ecoflex 00-30, Smooth-On). The base liquid and curing agents were mixed at a 1:1 mass ratio, thoroughly degassed in a vacuum chamber, and cast into resin molds printed by a 3D printer (Form 3, Formlabs). 
After the silicone rubber was fully cured, the cylindrical body and inlet component were removed from the resin molds and bonded together with uncured silicone rubber to ensure an airtight pneumatic seal. Then, a polyurethane tube for pneumatic pressure was inserted into the inlet component. The photograph of the fabricated cylindrical body is shown in Figure~\ref{fig3}(a).

 The external skin was designed using a stretchable textile. A high-stretch knit fabric (wt-10-1099150, Takisada Nagoya Co., Ltd.) was used for the actuator's side surfaces. This fabric exhibits mechanical anisotropy; thus, it was oriented so that its high-elasticity axis aligned with the actuator's longitudinal axis. To facilitate the attachment of constraint parts, a grid of buttons was sewn onto the skin using an electronic sewing machine (S71-SL, Brother Industries, Ltd.). To prevent unexpected deformation at the proximal and distal ends where constraint parts cannot be attached, these extremities were reinforced with a less-stretch woven fabric lining. The top portion of the actuator was capped with a similar low-stretch fabric. The constraint parts were fabricated from a less-stretchable satin fabric (Plain Satin Fabric, Marushin Co., Ltd.). The fabric was cut from a template into rectangular strips measuring $160\,\text{mm} \times 15\,\text{mm}$. Buttonholes, each 12 mm in length, were precisely opened at 24 mm intervals to ensure a secure and repeatable interface with the actuator’s button grid, shown in Figure~\ref{fig3}(b).
 For the twisting actuation, patch samples with buttonholes spaced at 33 mm intervals were additionally prepared.

 Figure~\ref{fig3}(c) shows the fabricated soft pneumatic actuator. The fabricated soft pneumatic actuator is cylindrical, with a total length of 160 mm and an outer diameter of 40 mm. The internal silicone bladder has a wall thickness of 5 mm. The arrangement of the constraint parts can be freely adjusted, as shown in Fig.~\ref{fig3}(d)-(f). These pictures show arrangements for bending, contraction, and twisting, respectively. The reconfiguration requires only a few tens of seconds without special tools. 

\subsection{Design Principle of Capacitive fabric-based tactile sensor}
\begin{figure}[t!]
    \centering
    \includegraphics[width=1.0\columnwidth]{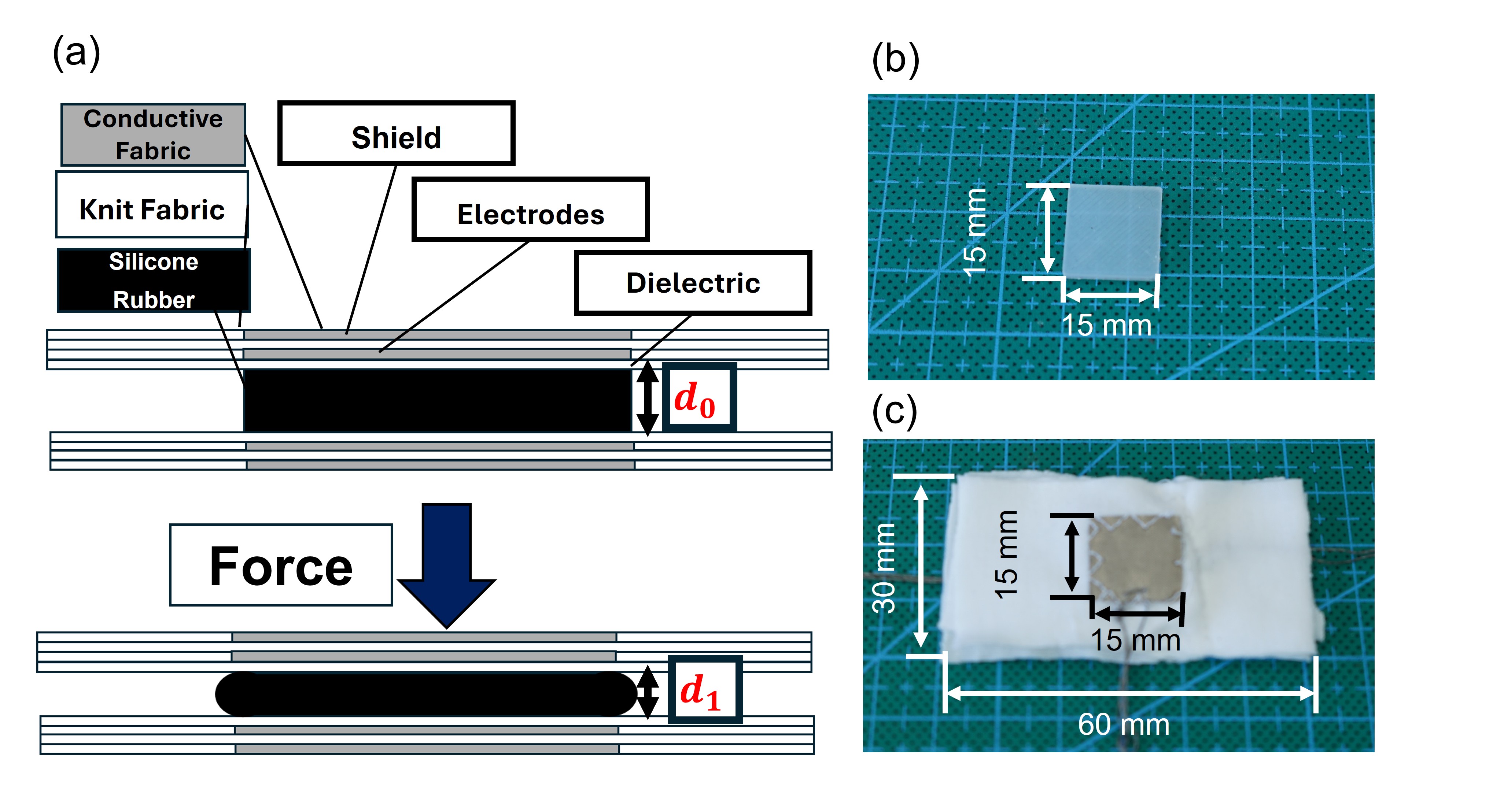} 
    \caption{Fabric-based tactile sensor and its principle. (a) Sensor structure and principle. (b) Overall view of the tactile sensor. (c) Silicone rubber was used as a dielectric layer. }
    \label{fig4}
\end{figure}
The proposed capacitive fabric force sensor detects external loads by measuring changes in capacitance caused by the deformation of a dielectric layer~\cite{b47, b48}. The tactile sensor is composed of conductive fabric electrodes sewn onto a fabric, with soft silicone rubber  acting as the dielectric layer.

The principle of the fabric-based tactile sensor is shown in Fig.~\ref{fig4}(a). This tactile sensor can also be fixed onto the soft actuator via buttons, in the same manner as the constraint parts. There are four fabric electrodes. The external two electrodes are for noise reduction, and the internal two electrodes are for capacitance measurement. The sensor can be regarded as a capacitor composed of two fabric layers and a silicone rubber layer.
When a vertical force is applied to the tactile sensor, the distance between the electrodes decreases, increasing capacitance. The total capacitance can be expressed by the following equation.
\begin{equation} \frac{1}{C_0} = \frac{1}{C_{f1}} + \frac{1}{C_{s}} + \frac{1}{C_{f2}} \end{equation}
Where:
\begin{description}
\item[\textbullet~$C_0$] :Total capacitance
\item[\textbullet~$C_{f1}$] :Capacitance of the top fabric layer
\item[\textbullet~$C_s$] :Capacitance of the silicone rubber layer
\item[\textbullet~$C_{f2}$] :Capacitance of the bottom fabric layer
\end{description}
 Capacitance of the fabric layers and the silicone rubber layer can be expressed by the following equation.
\begin{equation}
C_{f1} = C_{f2} = \frac{\varepsilon_0 \varepsilon_f A}{d_f}
\end{equation}
\begin{equation}
C_s = \frac{\varepsilon_0 \varepsilon_s A}{d_0}
\end{equation}
Where:
\begin{description}
\item [\textbullet~$\varepsilon_0$]: Permittivity of vacuum
\item [\textbullet~$A$]: Effective area of the fabric electrode
\item [\textbullet~$d_f$]: Thickness of the single fabric layer
\item [\textbullet~$\varepsilon_f$]: Relative permittivity of the fabric layer
\item [\textbullet~$d_0$]: Thickness of the silicone rubber
\item [\textbullet~$\varepsilon_s$]: Relative permittivity of the silicone rubber
\end{description}

Using these three equations, $C_0$ can be expressed as follows.
\begin{equation} C_0 = \frac{\varepsilon_0 A}{\frac{2d_f}{\varepsilon_f} + \frac{d_0}{\varepsilon_s}} \end{equation} When an external force is applied, the dielectric layer is compressed, reducing the thickness of the silicone rubber from its initial value $d_0$ to $d_1$.
\begin{equation} C_1 = \frac{\varepsilon_0 A}{\frac{2d_f}{\varepsilon_f} + \frac{d_1}{\varepsilon_s}} \end{equation}
Using the capacitance after deformation $C_1$, the capacitance change ratio can be expressed as follows: 
\begin{equation}
\begin{aligned}
\frac{\Delta C}{C_0}
&= \frac{C_1 - C_0}{C_0} \\
&= \frac{\varepsilon_f (d_0 - d_1)}
       {2\varepsilon_s d_f + \varepsilon_f d_0}
\end{aligned}
\end{equation}
Since the silicone layer deforms more than the fabrics due to its lower Young's modulus, the change in $d_0$ primarily governs the overall capacitance increase. We use this equation to calculate the theoretical value of the capacitance change ratio.

\subsection{Fabrication Process of Capacitive fabric-based tactile sensor}
The capacitive fabric-based tactile sensor was fabricated by layering conductive fabrics and a soft silicone rubber to ensure both structural flexibility and ease of assembly. The detailed structure of the sensor is depicted in Fig.~\ref{fig4}(a). For the fabric electrodes, a conductive adhesive fabric tape (Faraday Cloth Tape, uxcell) was utilized to simplify the fabrication process. Both the internal sensing electrodes and the external shield electrode were formed into $15\,\text{mm} \times 15\,\text{mm}$. The shield electrode was integrated to cover the sensing area, reducing  external noises and parasitic capacitance. For the dielectric layer, a soft silicone rubber (Ecoflex 00-30, Smooth-On) was used. The silicone rubber was cured into the size of $15\,\text{mm} \times 15\,\text{mm} \times 3\,\text{mm}$ using a resin mold printed by a 3D printer (Bambu Lab P1S Combo, Bambu Lab), as shown in Fig.~\ref{fig4}(b). The fabrics and the dielectric layer were assembled by stacking the two conductive tape electrodes with the 3 mm thick silicone dielectric layer sandwiched between them, as shown in Fig.~\ref{fig4}(c). 
The specific design parameters of the sensor in this study were established as follows. The relative permittivity values for the silicone rubber, which serves as the dielectric layer, and the knit fabric were determined from previously reported measurements. The electrode area $A$ is 225 $\text{mm}^2$ (15 mm $\times$ 15 mm), the initial thickness of the silicone rubber $d_s$ is 3.0 mm with a relative permittivity $\varepsilon_s$ of 3.4 ~\cite{b53}, and the thickness of the fabric layer $d_f$ is 0.3 mm with a relative permittivity $\varepsilon_f$ of 1.5~\cite{b54}. 
\section{Experimental Procedure}
We describe the experimental setups and procedures for evaluating the soft-fabric-based pneumatic actuator and the capacitive fabric-based tactile sensor.
\subsection{Experiment for actuation}
\begin{figure}[b!]
    \centering
    \includegraphics[width=1.0\columnwidth]{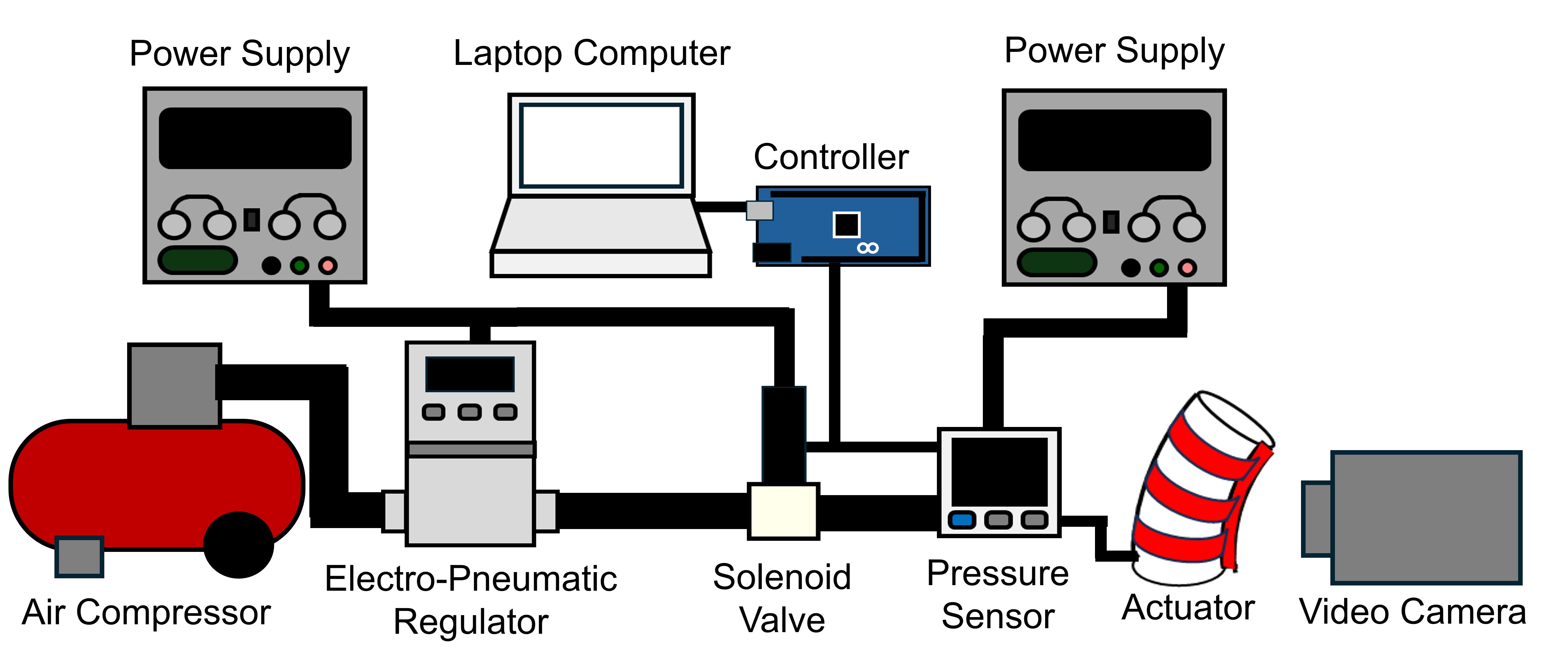} 
    \caption{Experimental setup to evaluate a fabric-based pneumatic actuator. The fabric-based pneumatic actuator was actuated by pneumatic pressure, and its motion was recorded with a video camera. The actuation was analyzed using image processing.}
    \label{fig5}
\end{figure}
We evaluated fabric-based pneumatic actuators to verify that the arrangement of the constraint parts can alter actuation. The schematic illustration of the experimental setup is shown in Fig.~\ref{fig5}.
The experimental setup consisted of a fabric-based pneumatic actuator, an air compressor (SR-102 SR-L30MPT-01, Fujiwara Sangyo), two electric power supplies (IPS303DD, ISO-TECH), a laptop computer (AERO5, GIGABYTE), a micro-controller (Arduino MEGA 2560 R3, Arduino), an electro-pneumatic regulator (EVD-1100-008AN, CKD), a solenoid valve (VQ110N-6LO, SMC), a pressure sensor (PPX-R01NH-6M, CKD) and a video camera (FDR-AX77, SONY).
During the evaluation, pneumatic pressure from the air compressor was supplied to the fabric-based pneumatic actuator. The pneumatic pressure was regulated using the electro-pneumatic regulator. The micro-controller was utilized to control the state of the solenoid valves. The fabric-based pneumatic actuator was mounted on a resin base to accurately record its shape.
Electric power was supplied to the electro-pneumatic regulator, the microcontroller, and the solenoid valve via power supplies.\par
The shape of the fabric-based pneumatic actuator induced by pneumatic pressure was recorded using the video camera at a frame rate of 30~fps and a resolution of $1920\,\text{pixel} \times 1080\,\text{pixel}$. 
Markers were used to track its shape changes. The marker detection in the recorded video was performed for each frame using Detectron2, an object detection framework~\cite{detectron2}. 
The centroid coordinates of the detected markers were extracted as the marker positions and tracked over time to analyze the deformation behavior. Based on the relative positions between the markers placed at the top and bottom of the actuator, we computed the elongation ratio, contraction ratio, bending angle, and twisting angle for each constraint pattern.

This evaluation focused on four actuation modes: no constraint, bending, twisting, and contracting. Based on preliminary trials, the maximum applied pressure for each mode was set to 24~kPa to avoid excessive inflation. The pressure was increased in increments of 3~kPa. Pneumatic pressure was applied and kept for 60~seconds to ensure the actuation reached a stable state. Five measurements were conducted for each pressure condition to evaluate actuation over time. To evaluate actuator-to-actuator variability, we fabricated three actuators and conducted evalutions for each actuator to analyze their individual differences.
After that, we evaluated the repeatability of each actuation and performed cyclic pressure-loading and unloading tests. A single cycle was defined as applying 24~kPa for 15~s followed by a 15~s depressurization period. This cycle was repeated 10 times, with the pressure states controlled via the solenoid valve. 

\subsection{Experiment for blocking force}
\begin{figure}[t!]
    \centering
    \includegraphics[width=1.0\columnwidth]{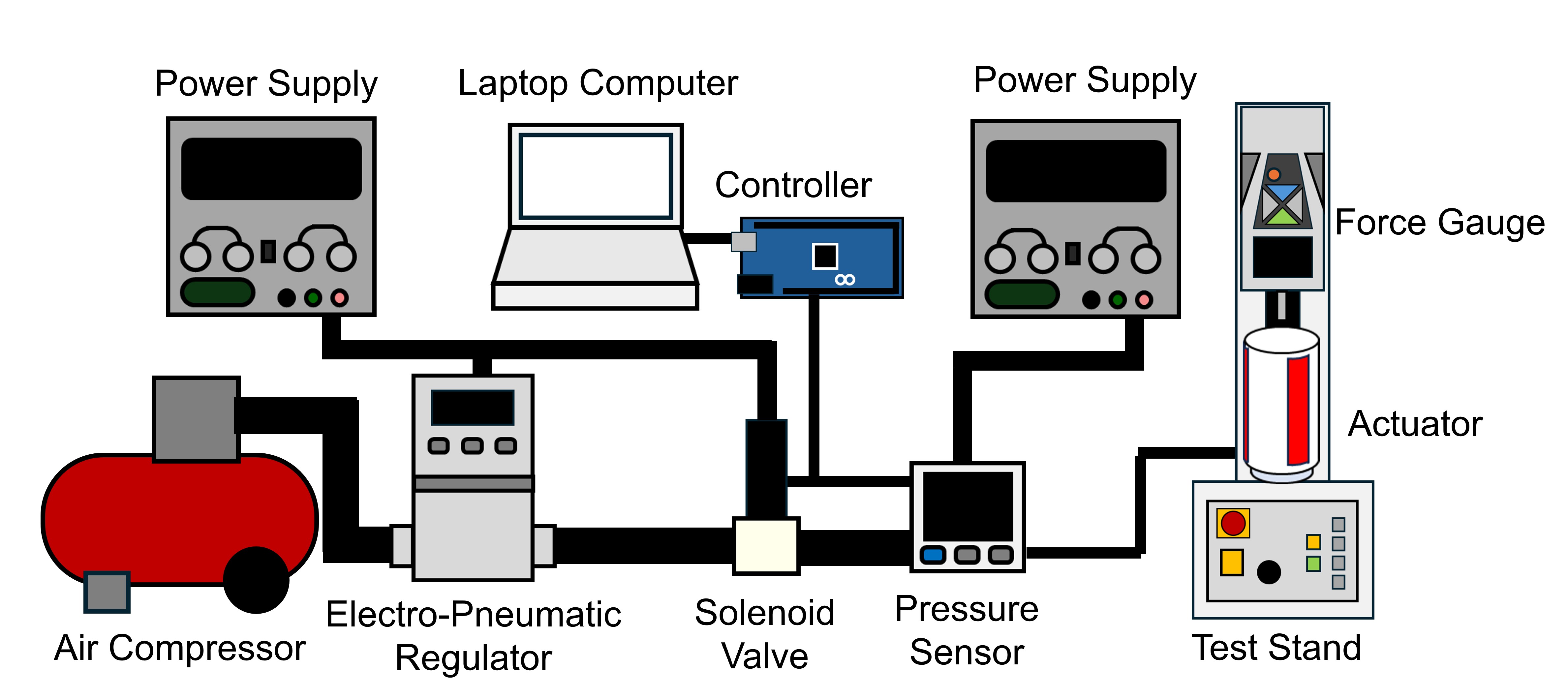} 
    \caption{Experimental setup to evaluate the blocking force of a soft pneumatic actuator. A test stand was used to fix the actuator, and the blocking force was measured using a force gauge.}
    \label{fig_blocking}
\end{figure}

We measured the blocking force generated by the fabric-based pneumatic actuator under load constraints. The blocking force was measured under each actuation mode. A schematic diagram of the experimental setup is shown in Fig.~\ref{fig_blocking}. The pneumatic pressure control system from the previous experiment was used for this one.
The fabric-based pneumatic actuator was mounted on a motorized test stand (MX-500N, Imada) using a resin fixture. The blocking force was measured using a force gauge (ZTS-50N, Imada). During the measurements, the actuator's deformation was completely constrained by the force gauge and the fixture. Pneumatic pressure was applied to the fabric-based pneumatic actuator, and the blocking force was recorded once stabilized. The applied pressure was increased from 0 to 24~kPa in increments of 3~kPa.  The blocking force was measured three times for each mode. 
\subsection{Experiment for fabric-based tactile sensor}
\begin{figure}[t!]
    \centering
    \includegraphics[width=1.0\columnwidth]{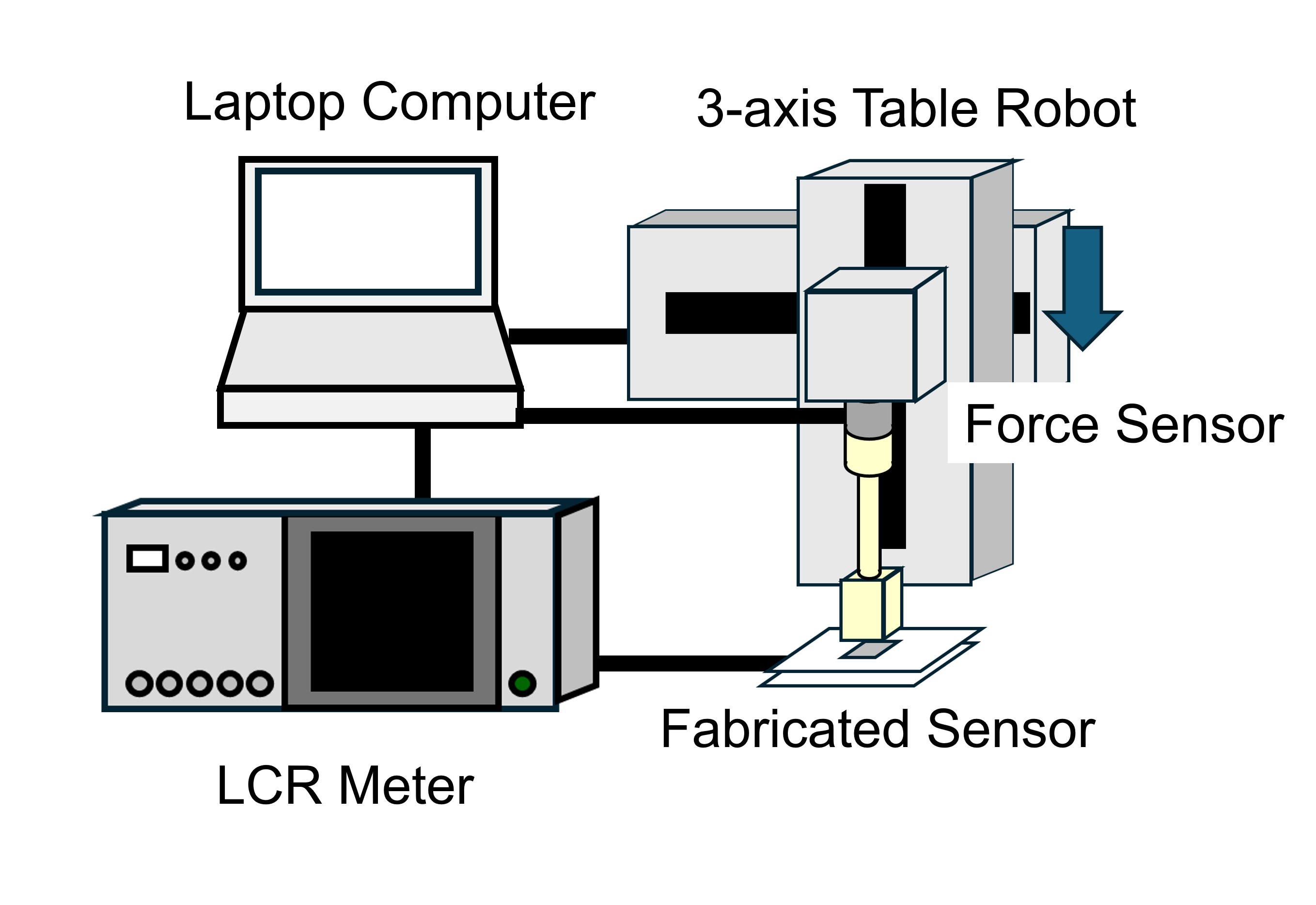} 
    \caption{Experimental setup to evaluate a capacitive fabric-based tactile sensor. A three-axis table robot was used to apply force to the fabric-based tactile sensor, and the sensor output was measured using an LCR meter.}
    \label{fig6}
\end{figure}
We evaluated the characteristics of the capacitive tactile sensor. The experimental setup is shown in Fig.~\ref{fig6}. The fabricated fabric-based tactile sensor was fixed to a 3-axis table robot (ICSB3-BA1MB1L-WA-30AQ-25AQ-10AQB-T2-3L-CT-CT, IAI). 
The fabric-based tactile sensor was compressed by the table robot, and the reaction force was measured using a 6-axis force sensor (PFS020YA500U6, Leptrino) with a square probe with a contact area of $12\,\text{mm} \times 12\,\text{mm}$. The output of the fabric-based tactile sensor was measured by an LCR meter (IM3533, Hioki) and compared with the force measured by the 6-axis force sensor to obtain the relationship between applied force and output. The measurement frequency and the measurement voltage were 100~kHz and 0.5~V, respectively. The measured output was recorded on the laptop at a 100 ms sampling interval for detailed analysis. In the experiment, starting from a state in which the probe was in contact with the sensor, the probe was pressed downward at a vertical speed of 0.1 mm/s to a maximum indentation depth of 1 mm. A pre-load of 2.5~N was applied to fill the gaps between the layers. The contact force and sensor output were recorded during this process. The initial capacitance  $C_0$ under no-load conditions was subtracted from the measured output to calculate the capacitance change ratio $\Delta C / C_0$. 
The hysteresis characteristics of the fabric-based tactile sensor were also evaluated by loading and unloading cycles.
Additionally, to evaluate the repeatability of the fabric-based tactile sensor, a cyclic loading test was performed. 10 continuous cycles of loading and unloading at a constant speed of 0.1~mm/s were applied to the fabric-based tactile sensor. The maximum indentation depth was 1~mm. The capacitance was recorded throughout the cycles.

\section{Experimental Results}
\begin{figure*}[t!] 
    \centering
    \includegraphics[width=\textwidth]{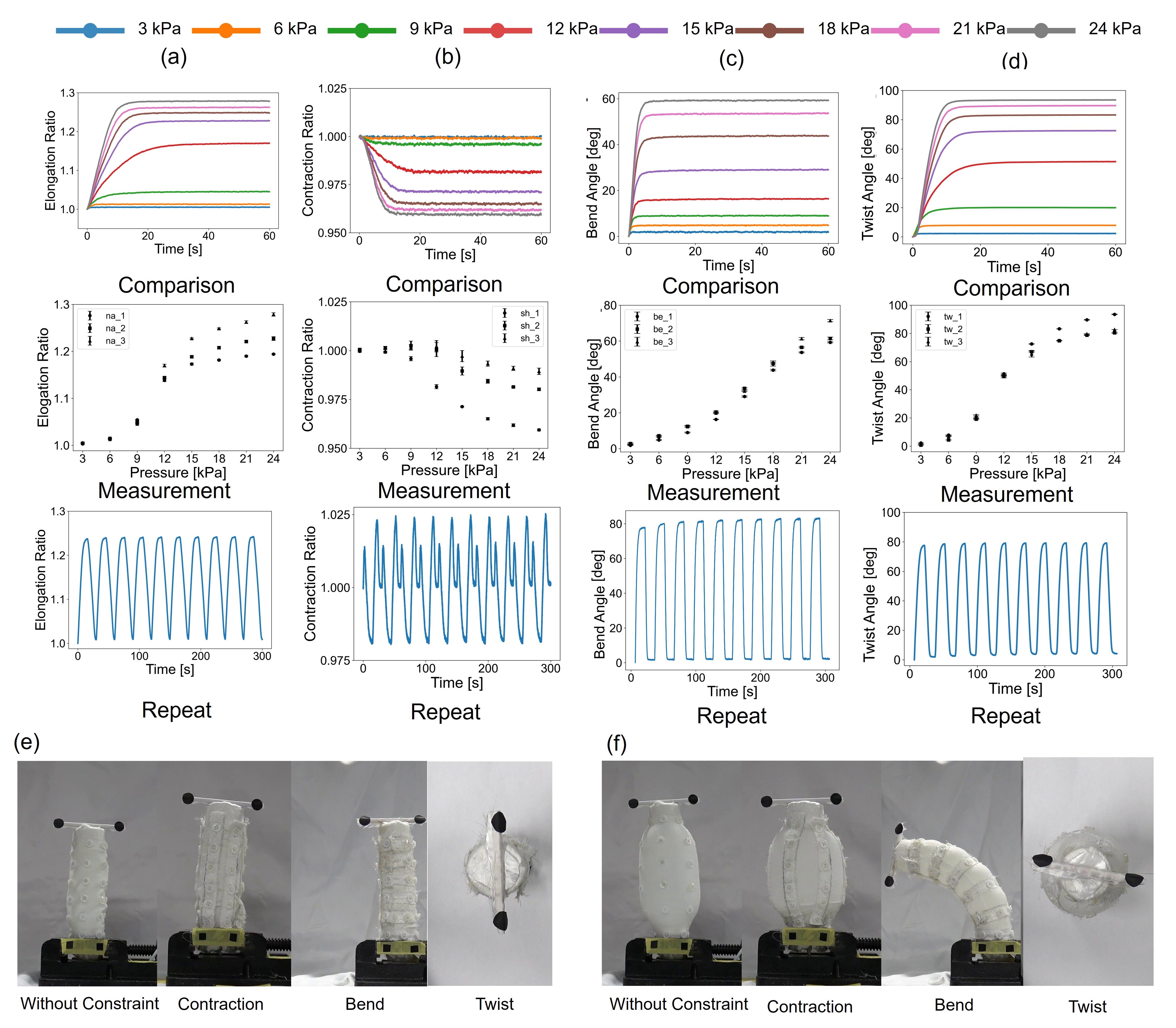}
    \captionsetup{justification=justified}
    \caption{Fabric-based pneumatic actuators under constraints actuated as designed. (a) Without constraints, the fabric-based pneumatic actuators expanded in all directions. Approximate increase in elongation ratio in the longitudinal direction was 20\%. (b)The fabric-based pneumatic actuators contracted to approximately 96\% of the initial length. (c) The fabric-based pneumatic actuators performed bending actuation,  approximately 60~degrees. (d) Under helical constraints, the fabric-based pneumatic actuators achieved a twist angle of 90~degrees. (e),(f) Experimental photographs: The left and right images show the initial state and the state when maximum pressure was applied, respectively.}
    \label{fig:fig7}
\end{figure*}
\subsection{EXPERIMENT FOR ACTUATION}
As illustrated in Fig.~\ref{fig:fig7}, each constraint pattern produced the intended deformation mode. These results demonstrate the effectiveness of the proposed method, which modifies actuation by simply applying localized constraint parts. Representative shape changes of the fabric-based pneumatic actuator are shown in Fig.~\ref{fig:fig7}(e) and (f). In the following, we describe the results for each constraint pattern. The top part shows the average and standard deviation over five trials. The middle part shows the stable values from five trials and its standard deviation.

In the no-constraint condition (baseline), the fabric-based pneumatic actuator exhibited approximately isotropic expansion. To quantify axial deformation, we calculated the elongation ratio from the change in vertical displacement. The top part of Fig.~\ref{fig:fig7}(a) shows the elongation ratio over time. The elongation ratio increased over time and finally saturated. The maximum elongation ratio tended to saturate above 12~kPa. As pressure increases, the tensile load applied to the fabric skin also increases; however, under high tensile loads, the fabric becomes more difficult to stretch further due to tension-induced stiffening. Consequently, at higher pressures, the elongation of the fabric skin was suppressed, and the actuator exhibited saturated axial extension. At 24~kPa, the actuator reached an elongation ratio of 1.2 (i.e., 120\% of its initial length).
It was confirmed that the relationships between pneumatic pressure and maximum elongation ratio for the three fabric-based pneumatic actuators did not perfectly coincide, attributed to fabrication-induced differences, as shown in the middle part of Fig.~\ref{fig:fig7}(a). Nevertheless, the maximum difference was within 10\%, and it should be possible to further reduce this error by refining the fabrication process. Owing to the actuator’s low viscoelasticity, no change in the maximum elongation ratio was observed during loading and unloading cycles, as shown in the bottom part of Fig.~\ref{fig:fig7}(a).

The contraction ratio, defined as the axial displacement normalized by the initial length, decreased over time, as shown in the top part of Fig.~\ref{fig:fig7}(b). 
The trend indicates that the fabric-based actuator contracted as we designed. 
Similar to the no-constraint condition, the contraction ratio was saturated at higher pneumatic pressure, which can be attributed to the increased effective stiffness of the fabric under high tensile load (tension-induced stiffening). A comparison among the three fabricated actuators revealed some variability in the contraction ratio, likely due to fabrication tolerances (middle part of Fig.~\ref{fig:fig7}(b)); however, the differences were limited because the overall contraction ratio was relatively small. During repeated actuation cycles, the contraction ratio remained unchanged, indicating stable repeatability (bottom part of Fig.~\ref{fig:fig7}(b)).
In some trials, a slight initial extension phase prior to contraction was observed. This behavior may depend on microscopic initial gaps and the contact conditions between the internal bladder and the outer fabric/constraint parts. Notably, this initial extension was not consistently observed across actuators (see the top part of Fig.~\ref{fig:fig7}(a)), suggesting that it is sensitive to fabrication- and assembly-dependent conditions.

As shown in Fig.~\ref{fig:fig7}(c), the bending constraint pattern successfully induced a bending actuation. The maximum bending angle reached approximately 60~degrees at 24~kPa and increased approximately linearly with pneumatic pressure, as shown in the top and middle parts of Fig.~\ref{fig:fig7}(c). This trend differs from those observed in Fig.~\ref{fig:fig7}(a) and (b), which is likely attributable to the anisotropic stretchability of the fabric material. The variation among the three actuators was small. During loading and unloading cycles, a slight change in the bending angle was also observed, which may be caused by the same fabric anisotropy, as shown in the bottom part of Fig.~\ref{fig:fig7}(c).

The twisting constraint pattern induced twisting actuation, as shown in Fig.~\ref{fig:fig7}(d). Upon pressurization, the helical constraint parts effectively converted the expansion energy into a rotational actuation. At the maximum pressure of 24~kPa, the fabric-based actuator exhibited a significant twisting angle of approximately 90 degrees, as shown in the top part of Fig.~\ref{fig:fig7}(d). 
The twisting angle increased with pressure, but tended to saturate above 18~kPa, as shown in the middle part of Fig.~\ref{fig:fig7}(d). The variation among the three actuators was small.
During loading and unloading cycles, no change in the twisting angle was also observed, and we confirmed the repeatability of the fabric-based actuator with the constraint pattern, as shown in the bottom part of Fig.~\ref{fig:fig7}(d).
\begin{figure}
    \centering
    \includegraphics[width=1.0\columnwidth]{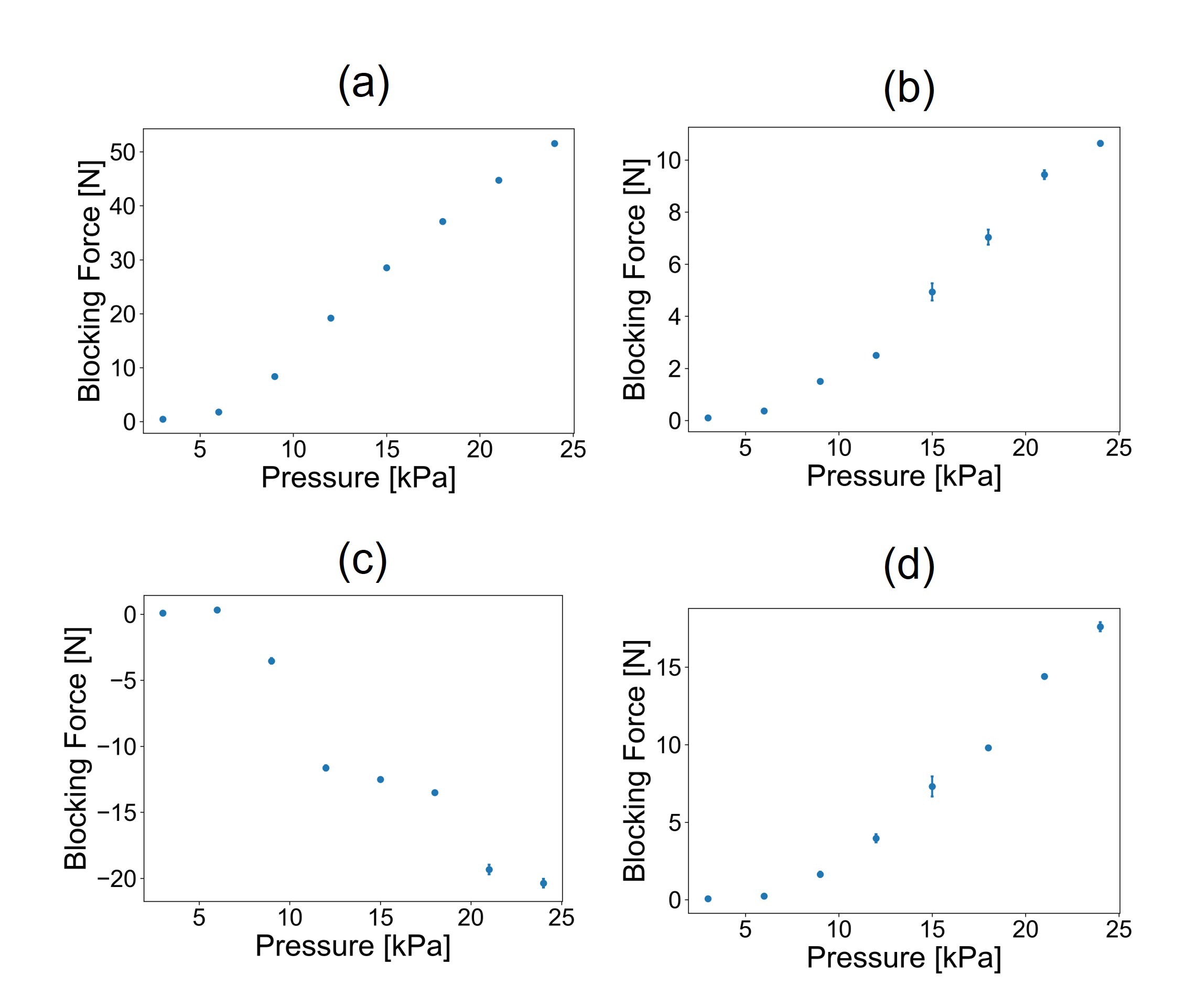} 
    \caption{Blocking force for each constraint pattern increased with the applied pneumatic pressure. At low pressures, the blocking force remained small, likely because most of the pneumatic energy was consumed in expanding the actuator rather than being converted into external force. (a) No constraint. (b) Contraction. (c) Bending. (d) Twisting.}
    \label{fig_result_blocking}
\end{figure}
\begin{figure}[b!]
    \centering
    \includegraphics[width=\columnwidth]{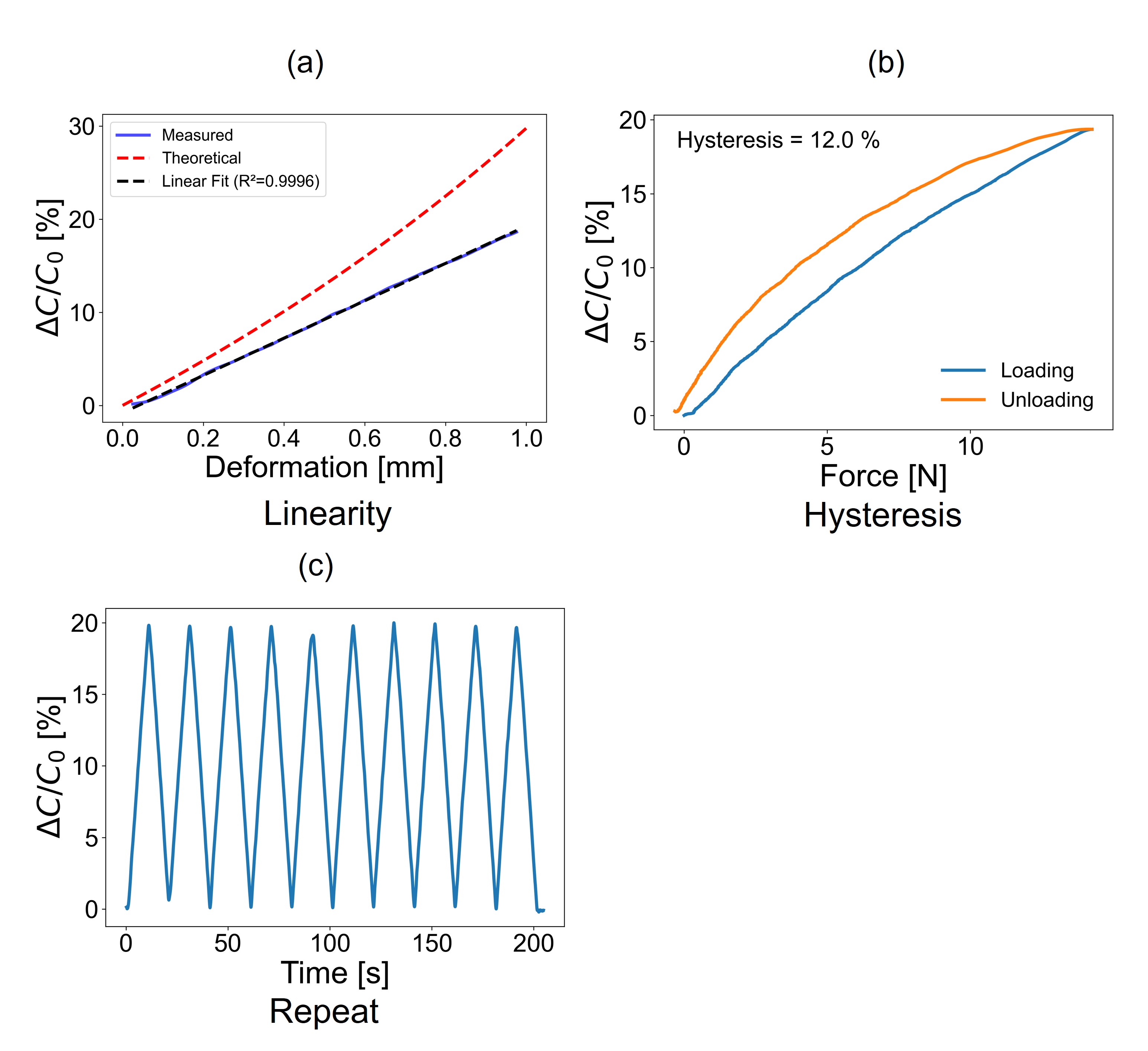} 
    \captionsetup{justification=justified}
    \caption{The soft force sensor's output increased with increasing force, as designed. (a), (b), and (c) show the results for the tactile sensor we fabricated. High linearity and reproducibility were confirmed.}
    \label{fig:fig8}
\end{figure}

\subsection{EXPERIMENT FOR BLOCKING FORCE}
Figure~\ref{fig_result_blocking} shows the measured blocking-force results. In this study, we evaluated the maximum force the actuator can produce. The figure shows the maximum values from three trials and its standard deviation. Fig.~\ref{fig_result_blocking}(a) presents the pushing force generated during actuator elongation without constraint parts, Fig.~\ref{fig_result_blocking}(b) presents the tensile force generated during contraction, Fig.~\ref{fig_result_blocking}(c) presents the vertical pushing force produced by bending, and Fig.~\ref{fig_result_blocking}(d) presents the pushing force generated by twisting. The experimental results indicate that the generated force increased with pneumatic pressure, and that in all cases a force of more than 10~N was achieved at an applied pressure of 24~kPa.

\subsection{EXPERIMENT FOR fabric-based tactile sensor}
Figure~\ref{fig:fig8} shows the experimental results for the capacitive fabric-based tactile sensor. The capacitance change ratio $\Delta C/C_0$ was increased with the increase in the applied force, as shown in Fig.~\ref{fig:fig8}(a). This trend accurately reflects the elastic deformation of the dielectric silicone layer under small compression. Although both the theoretical and experimental results exhibited a similar monotonic increase, a discrepancy of approximately 50\% was observed between them. This mismatch may be attributed to variations in the material properties and to insufficient modeling of the gaps between the fabric layers. The calculated hysteresis was approximately 12\%, demonstrating the small viscosity of the dielectric layer despite the small deviation, as shown in Fig.~\ref{fig:fig8}(b). Fig.~\ref{fig:fig8}(c) shows the sensor output during these loading and unloading cycles. No noticeable drift in the capacitance change was observed over repeated cycles.

These results represent an idealized case obtained by eliminating the influence of the tactile sensor’s layered structure, applying a pre-load. In practice, it is necessary to construct a calibration curve relating the sensor output to the applied force, while accounting for the layered structure. In the following, we therefore use the tactile sensor only for contact detection and do not use it for quantitative force measurement.
\section{Application}
\begin{figure}[t!]
    \centering
    \includegraphics[width=1.0\columnwidth]{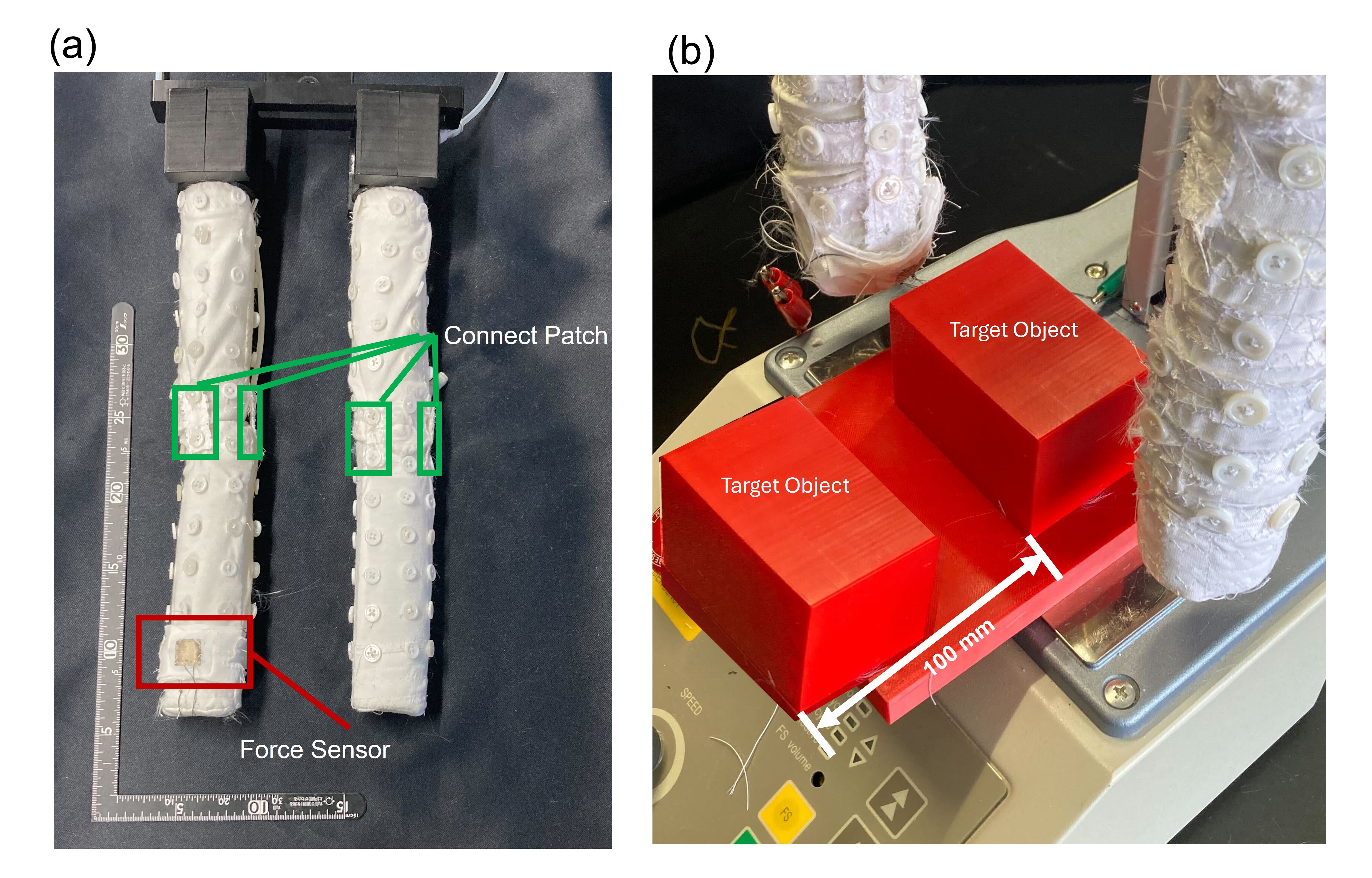} 
    \caption{As an application, we constructed two soft robotic arms and conducted object-grasping experiments. (a) Each soft robotic arm consisted of two soft pneumatic actuators. By altering the constraint pattern on the actuator surface, the motion of the soft robotic arms could be reconfigured. (b) We confirmed that reconfiguring the motion enabled adjustment of the grasping position.}
    \label{fig_robot_env}
\end{figure}

We fabricated a soft robotic arm by connecting the proposed fabric-based pneumatic actuators using  fabric connector parts. Figure~\ref{fig_robot_env}(a) shows the assembled soft robotic arm. Each arm consists of two fabric-based pneumatic actuators, and we used two such arms to grasp objects. Fabric-based tactile sensors were attatched at the tip of the soft robotic arm to confirm successful grasping via bottons on the fabric-based pnematic actuators. In this experiment, we demonstrate that the overall robot motion can be reconfigured by altering the actuator constraint patterns.

Figure~\ref{fig_robot_env}(b) indicates the positions of the target objects. In the first configuration, the upper actuator was set to the contraction mode, while the lower actuator was set to the bending mode. With this combination, the arm successfully grasped an object directly beneath it. Next, we changed the constraint patterns so that both the upper and lower actuators operated in the bending mode. This reconfigured motion was used to grasp an object farther from the arm.

\begin{figure*}[t!]
    \centering
    \includegraphics[width=0.9\textwidth]{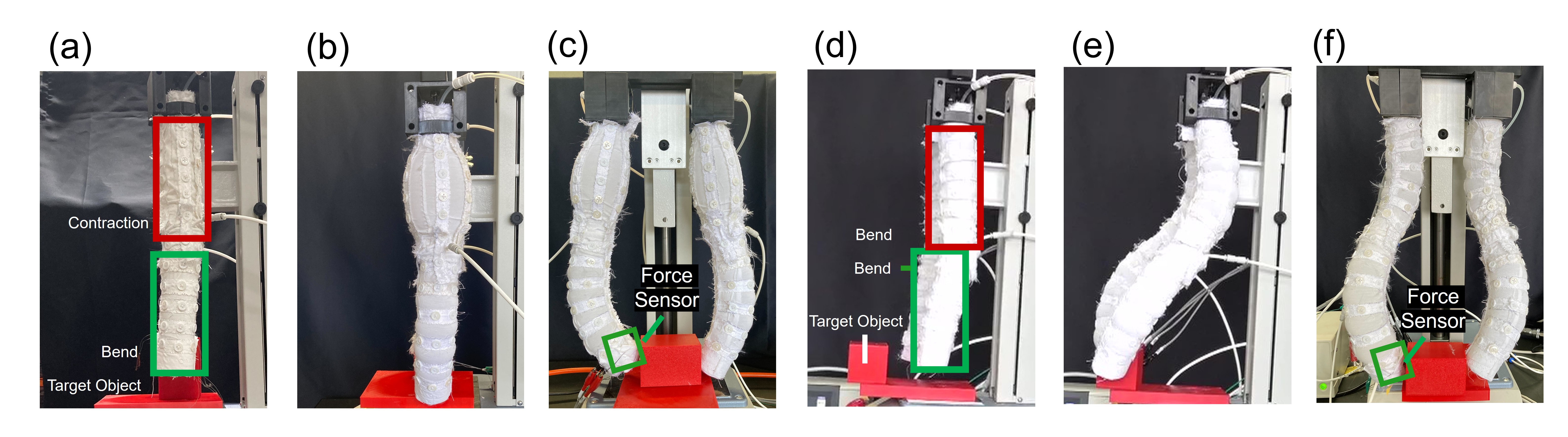} 
    \caption{By altering the constraint patterns of the soft robotic arms as designed, we were able to alter the grasping position. (a)–(c) The upper actuators were configured for contraction, while the lower actuators were configured for bending, enabling the soft robotic arms to grasp an object located directly beneath them. (d)–(f) By configuring all actuators for bending, the soft robotic arms were able to grasp an object located in front of the arm. These results demonstrate that the arm motion can be reconfigured solely by changing the surface constraint patterns, without disassembling the soft robotic arms.}
    \label{fig_robot_result}
\end{figure*}

\begin{figure}[t!]
    \centering
    \includegraphics[width=1.0\columnwidth]{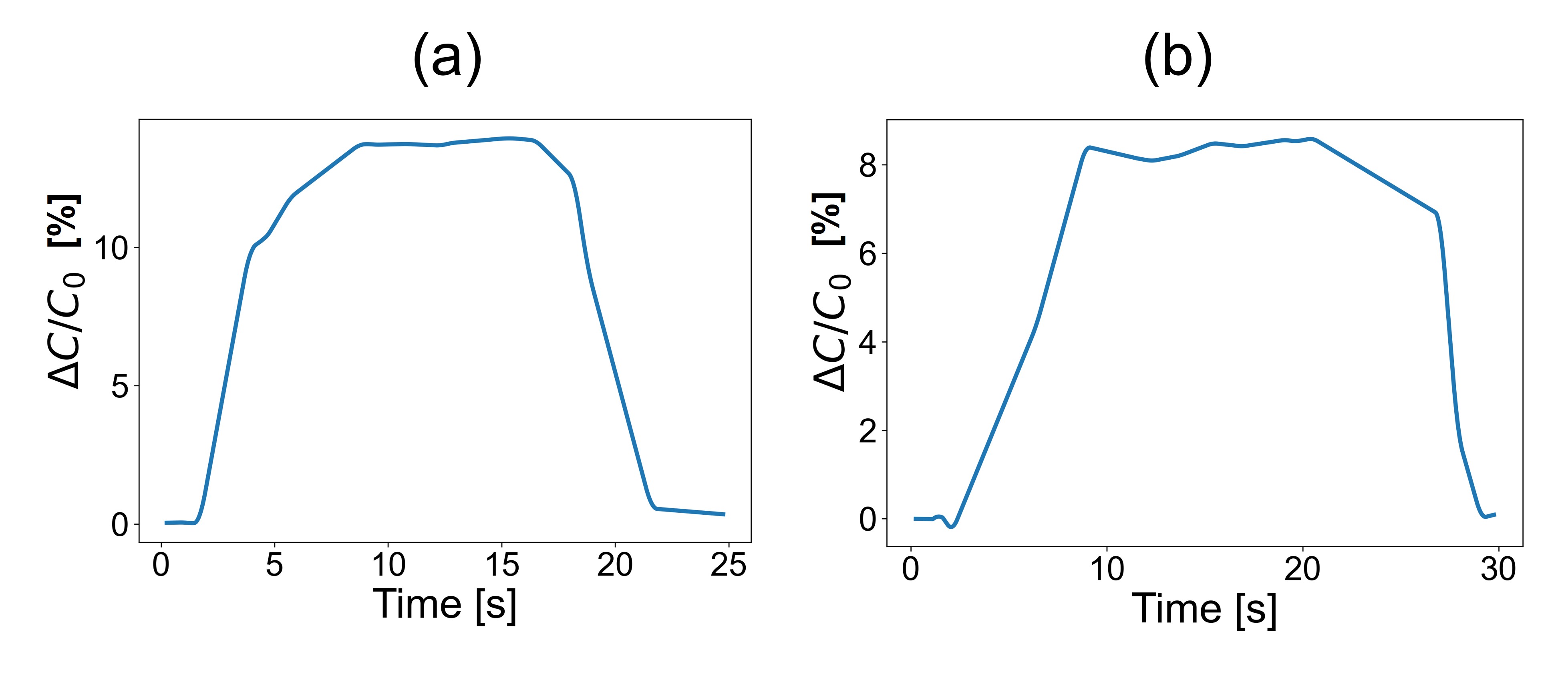} 
    \caption{Sensor outputs during object grasping with the soft robotic arms. (a) Sensor output when grasping an object located directly beneath the soft robotic arms. (b) Sensor output when grasping an object located in front of the soft robotic arms.}
    \label{fig_robot_sensor}
\end{figure}

The experimental results are shown in Fig.~\ref{fig_robot_result}. When combining a contracting actuator with a bending actuator, the soft robotic arm grasped the object positioned directly below (Fig.~\ref{fig_robot_result}(a)-(c)). The change in sensor output reflected the grasping state (Fig.~\ref{fig_robot_sensor}(a)). After reconfiguring the constraints, the arm motion changed accordingly, and the grasping of the object located in front of the arm was confirmed both visually (Fig.~\ref{fig_robot_result}(d)-(f)) and by the sensor response (Fig.~\ref{fig_robot_sensor}(b)).

These results indicate that implementing the proposed soft pneumatic actuators as a robotic arm enables a soft robot whose behavior can be reconfigured simply. While previous studies have reported that the grasping behavior of a soft robot can be changed by replacing soft pneumatic actuator modules~\cite{b52}, the proposed approach allows the grasping motion to be modified without disassembling the soft robotic arm.

\section{discussion}
Using the proposed method, we confirmed that the motion of a fabric-based pneumatic actuator can be reconfigured by attaching detachable, low-stretch fabric parts. Since the reconfiguration only requires attaching inextensible fabric parts, the actuation mode can be changed more simply and in less time than in related approaches. However, practical issues remain regarding the materials' stretchability and mechanical strength. If the textile used as the external skin could be designed in-house, it may be possible to obtain an application-optimized skin. In particular, the maximum allowable pressure is currently limited by the silicone rubber and knitted fabric, and thus, the actuator can be driven only at a lower pressure than other textile-based pneumatic actuators~\cite{b43, b44}. If this limitation can be mitigated, both the achievable displacement and the output force should be improved.

Although this study focused on soft pneumatic actuators, the proposed approach can be broadly applied to soft pneumatic robots composed of fabrics~\cite{b49, b50}. For example, it could be extended to humanoid inflatable robots, enabling their motions to be modified by attaching inextensible materials such as fabrics or tapes. In this work, we also proposed sensing components and demonstrated their usefulness through implementation in a robot; similarly, strain sensors could be realized by combining stretchable materials with soft conductive materials~\cite{b51}. In addition, we constructed a soft robot by connecting multiple actuators. Using a similar button-based connection structure, it should be possible to connect fabric-based actuators to achieve a variety of robot morphologies.

In this study, the soft pneumatic actuator was designed within the constraints of manual fabrication. Consequently, the number of constraints that could be placed was limited. By automating the fabrication process, it should be possible to attach buttons with a higher spatial density, enabling more complex motions by combining the basic motions investigated in this work. Automation would also help reduce the fabrication errors observed in the experimental results. For simple actuation such as bending and twisting, constraint patterns can likely be designed by referring to existing structures~\cite{b39, b40, b44}. However, for complex actuation, tools for designing the actuation itself will be required. This may further necessitate the development of computational methods, such as topology optimization~\cite {b41, b42}.
The actuator structure in this work consists of a thin fabric layer and a thick silicone rubber layer. The large thickness difference between the materials makes it difficult to generate appropriate finite element meshes, and thus the behavior cannot be easily computed using standard finite element methods. Therefore, to build a computational model, either an approximate actuator model must be developed, or the actuator structure must be reconsidered.
\section{conclusion}
In this paper, we proposed a reconfigurable fabric-based pneumatic actuator system with reusable, button-fastened constraint modules and an attachable capacitive textile tactile sensor. By selectively attaching low-stretch constraint parts to a fabric skin, a single actuator body achieved four distinct deformation modes—isotropic expansion, contraction, bending, and twisting — without modifying the internal chamber design or embedding reinforcements. Experimental results verified multi-mode actuation up to 24~kPa, including an elongation ratio of 1.2, a bending angle of approximately 60~degrees, and a twisting angle of approximately 90~degrees. Blocking-force measurements showed that forces exceeding 10~N were generated at 24~kPa across all modes. The capacitive fabric-based tactile sensor exhibited monotonic force-dependent output with low hysteresis (approximately 12\%) and stable responses over repeated loading cycles, supporting its use for contact detection on the actuator surface. We also demonstrated a soft robotic arm composed of multiple actuators and confirmed that its grasping motion can be reconfigured simply by changing the attached constraint patterns. Future work will focus on improving the allowable pressure and output performance through material and fabrication refinements, increasing the spatial density of the fastening interface for more complex motions, and developing computational design methods for constraint-layout generation.

\begin{IEEEbiography}[{\includegraphics[width=1in,height=1.25in,clip,keepaspectratio]{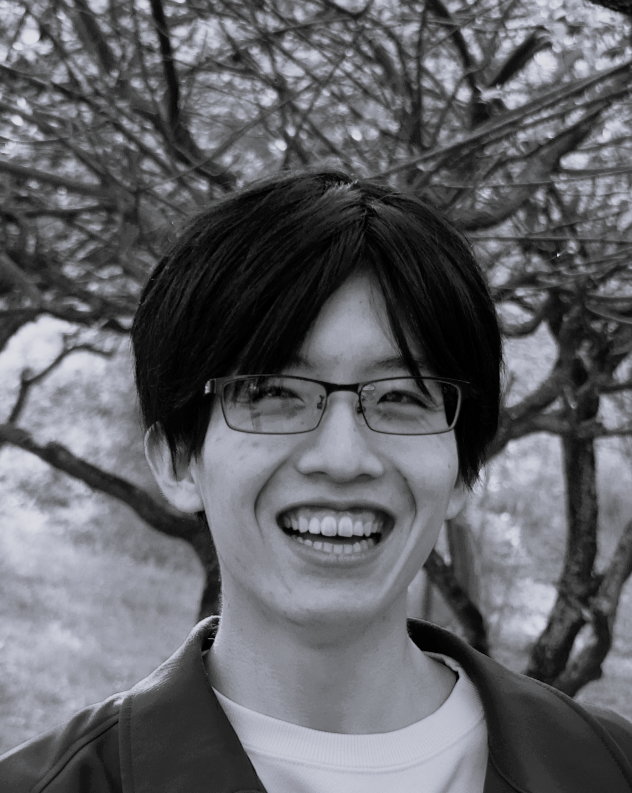}}]{Kentaro Kimura} received the B.E. degree from Osaka University, Toyonaka, Japan, in 2024. He is currently pursuing the master’s degree at the Graduate School of Engineering Science, Osaka University. His research interests include soft robotics and smart materials.
\end{IEEEbiography}
\begin{IEEEbiography}[{\includegraphics[width=1in,height=1.25in,clip,keepaspectratio]{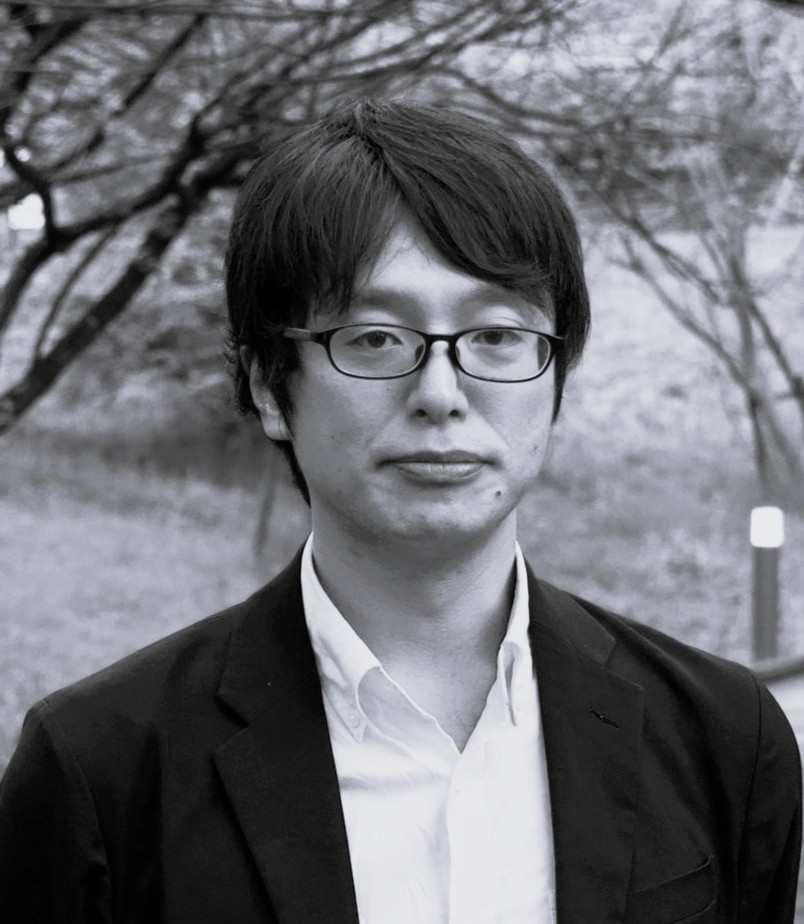}}]{Hiroki Ishizuka}
received the Ph.D. degree in integrated design engineering from Keio University, Japan, in 2016. He was an Assistant Professor with the Faculty of Engineering, Kagawa University, from 2016 to 2019. He is currently an Assistant Professor with the Graduate School of Engineering Science, Osaka University. His research interests include haptics and soft robotics.
\end{IEEEbiography}
\begin{IEEEbiography}[{\includegraphics[width=1in,height=1.25in,clip,keepaspectratio]{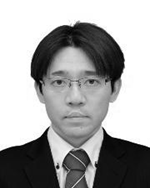}}]{Yusuke Sakaue}
eceived the B.S., M.E., and Ph.D. degrees from Ritsumeikan University, Shiga, Japan, in 2007, 2009, and 2015, respectively. He was a Postdoctoral Researcher at the Ritsumeikan Global Innovation Research Organization from 2015 to 2017. He worked in research and development for medical and healthcare equipment at Terumo Corporation from 2009 to 2012 and at Toray Engineering Co., Ltd., from 2017 to 2022. Since 2024, he has been a Lecturer in the Graduate School of Engineering Science, Osaka University. His research activities focus on biomedical engineering, covering a broad range from fundamental studies to practical applications. He is a member of the Japan Society for Medical and Biological Engineering and the Society of Instrument and Control Engineers.
\end{IEEEbiography}
\begin{IEEEbiography}[{\includegraphics[width=1in,height=1.25in,clip,keepaspectratio]{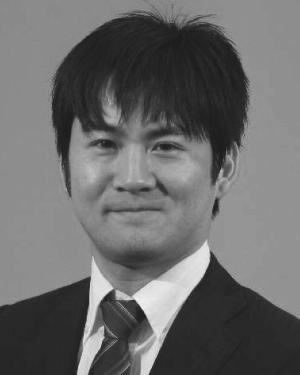}}]{Sei Ikeda}
received the Ph.D. degree from Nara Institute of Science and Technology, Ikoma, Japan, in 2006. He was an Associate Professor with the Graduate School of Engineering Science, Osaka University. He is currently a Professor with the Faculty of Advanced Science and Technology, Ryukoku University. His research interests include human–computer interaction, mixed reality, and computer vision.
\end{IEEEbiography}
\begin{IEEEbiography}[{\includegraphics[width=1in,height=1.25in,clip,keepaspectratio]{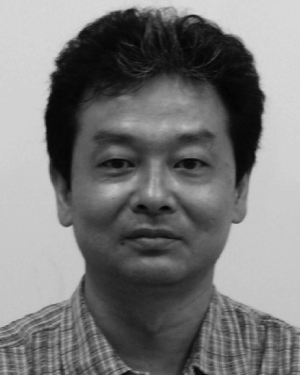}}]{Osamu Oshiro}
received the Ph.D. degree in engineering from Osaka University, Toyonaka, Japan, in 1990. He was a Research Engineer with Sumitomo Metal Industries from 1990 to 1993 and an Assistant/Associate Professor with Nara Institute of Science and Technology from 1993 to 2003. Since 2003, he has been a Professor with the Graduate School of Engineering Science, Osaka University. His research interests include PBR, facing technology, and modeling of distributed human systems.
\end{IEEEbiography}

\EOD

\end{document}